\documentclass[sigconf]{acmart}

\AtBeginDocument{%
  }

\usepackage{graphicx}
\usepackage{multirow} 
\usepackage{mathrsfs}
\usepackage{amsmath}
\usepackage{bm}     
\usepackage{makecell}
\usepackage{color}
\usepackage{amsfonts} 
\usepackage{float}
\usepackage{subfig}
\usepackage{balance}

\usepackage{pifont}
\newcommand{\cmark}{\ding{51}}

\newcommand{\raisedth}[1]{#1^{\textrm{th}}}
\newcommand{\bracks}[1]{\left[#1\right]}
\newcommand{\lfq}[1]{{\color{black}{#1}}}

\copyrightyear{2024}
\acmYear{2024}
\setcopyright{acmlicensed}
\acmConference[MM '24]{}{October 28-November 1, 2024}{Melbourne, VIC, Australia}
\acmBooktitle{Proceedings of the 32nd ACM International Conference on Multimedia (MM '24), October 28-November 1, 2024, Melbourne, VIC, Australia}
\acmDOI{10.1145/3664647.3680892}
\acmISBN{979-8-4007-0686-8/24/10}

\begin{document}

\title{Emphasizing Semantic Consistency of Salient Posture for Speech-Driven Gesture Generation}

\author{Fengqi Liu}
\orcid{0000-0002-1919-2167}
\authornote{Equal contribution.}
\affiliation{%
  \institution{Shanghai Jiao Tong University}
  \city{Shanghai}
  \country{China}
}
\email{liufengqi@sjtu.edu.cn}

\author{Hexiang Wang}
\authornotemark[1]
\affiliation{%
  \institution{Shanghai Jiao Tong University}
  \city{Shanghai}
  \country{China}
}
\email{whxsjtu123@sjtu.edu.cn}

\author{Jingyu Gong}
\authornotemark[1]
\affiliation{%
  \institution{East China Normal University}
  \city{Shanghai}
  \country{China}
}
\email{jygong@cs.ecnu.edu.cn}

\author{Ran Yi}
% \authornotemark[2]
\affiliation{%
  \institution{Shanghai Jiao Tong University}
  \city{Shanghai}
  \country{China}
  }
\email{ranyi@sjtu.edu.cn}

\author{Qianyu Zhou}
\affiliation{%
  \institution{Shanghai Jiao Tong University}
  \city{Shanghai}
  \country{China}
}
\email{zhouqianyu@sjtu.edu.cn}

\author{Xuequan Lu}
\affiliation{%
  \institution{La Trobe University}
  \city{Melbourne}
  \country{Australia}
}
\email{b.lu@latrobe.edu.au}

\author{Jiangbo Lu}
\affiliation{%
  \institution{SmartMore Corporation}
  \city{Shenzhen}
  \country{China}
}
\email{jiangbo.lu@gmail.com}

\author{Lizhuang Ma}
\authornote{Corresponding author.}
\affiliation{%
  \institution{Shanghai Jiao Tong University}
  \city{Shanghai}
  \country{China}
}
\email{ma-lz@cs.sjtu.edu.cn}

\renewcommand{\shortauthors}{Fengqi Liu et al.}

\begin{abstract}
Speech-driven gesture generation aims at synthesizing a gesture sequence synchronized with the input speech signal. Previous methods leverage neural networks to directly map a compact audio representation to the gesture sequence, ignoring the semantic association of different modalities and failing to deal with salient gestures. In this paper, we propose a novel speech-driven gesture generation method by emphasizing the semantic consistency of salient posture. Specifically, we first learn a joint manifold space for the individual representation of audio and body pose to exploit the inherent semantic association between two modalities, and propose to enforce semantic consistency via a consistency loss. Furthermore, we emphasize the semantic consistency of salient postures by introducing a weakly-supervised detector to identify salient postures, and reweighting the consistency loss to focus more on learning the correspondence between salient postures and the high-level semantics of speech content. In addition, we propose to extract audio features dedicated to facial expression and body gesture separately, and design separate branches for face and body gesture synthesis. Extensive experimental results demonstrate the superiority of our method over the state-of-the-art approaches. 
\end{abstract}




\maketitle

\section{Introduction}

\begin{figure}[t!]
    \centering
    \includegraphics[width=\linewidth]{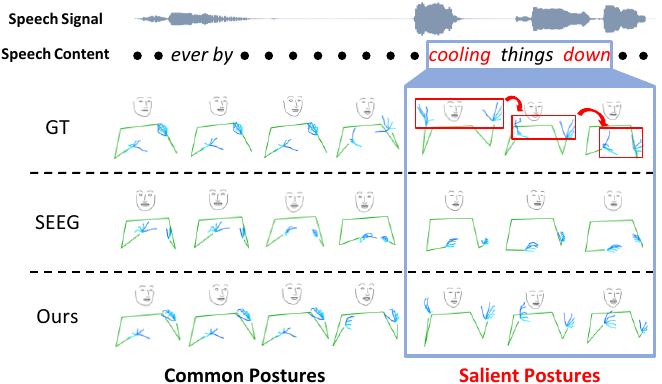}
    \caption{Salient postures indicate the large pose movements associated with the high-level semantics of speech content, e.g., \textit{cooling down}, which are hard to be generated. Our method can synthesize more realistic gestures than SEEG~\cite{liang2022seeg} by emphasizing semantic consistency of salient postures.}
    \label{fig:salient_posture}
\end{figure}

The task of speech-driven gesture generation aims to synthesize a sequence of gestures line with the given speech signal, which has a wide range of application scenarios, including online service~\cite{li2016social,liao2019embodying}, virtual avatar animation~\cite{zhuang2022text,liang2022seeg} and human-machine interaction~\cite{salem2011friendly,salem2012generation,carfi2021gesture}. Compared with lip motion generation, speech-driven gestures are more implicit and metaphoric, which makes the gesture generation task a non-trivial challenge.

Considering the significant modality gap between speech and gestures, traditional works~\cite{kopp2006towards,wagner2014gesture,cassell1994animated} tackle the problem through rule-based generation approaches, which establish the deterministic correspondence between audio syllables and gesture sets. Such methods ignore the intrinsic connection between different modalities and suffer from poor naturalness. 
Recent data-driven methods~\cite{liu2022learning,liang2022seeg} achieve better performance by utilizing deep neural networks to extract audio representation of different semantic granularities, which is then decoded to generate a holistic gesture sequence. 

However, these methods suffer from the following major weaknesses: 
(1) their pipelines are straightforward and cannot effectively achieve semantic consistency between the speech content and the synthesized gestures, especially for the postures with large movement scope.
(2) They treat facial expressions and body movements as a whole and simultaneously synthesize them with a single pipeline, which typically leads to poor synchronization between facial expressions and speech.

Therefore, our primary goal is to enhance the correspondence  between the generated gestures and the semantics in the speech content. From 
a human gesture study~\cite{freeman2011hand}, we identify one significant observation: postures with large movement scope in the sequence correspond to strong and rich semantic information of speech audio, and postures with slight movement scope correspond to weak semantics of speech audio. 
As illustrated in Figure~\ref{fig:salient_posture}, the gesture sequence of opening and raising the arms, then folding and lowering them is closely related to the phase \textit{cooling down}. For clarity, we define salient postures as the postures with large movement scope, i.e., the movements in which the shoulders and arms have a relatively large amplitude of motion, which are often related to strong semantics of the speech content. Salient postures in gesture sequences tend to be significant in conveying the intention and emotion of speakers, and therefore the exact correspondence relationship between salient postures and speech with strong semantics will contribute to the vivid and realistic gesture generation. 
% } 

Inspired by the above discussions, we propose a novel speech-driven gesture generation method by emphasizing semantic consistency of salient posture. To fully exploit the inherent semantic association between audio and gesture, we first learn a joint manifold space for the representations of audio and body pose to establish a mapping between the two modalities. 
%Specifically, we utilize separate feature extractors to extract the individual latent codes of audio and pose corresponding to the body part. 
Through the consistency loss, we ensure that the audio and pose features are close to each other in the shared joint-embedding space and represent similar semantic information. 

Furthermore, we propose to {\bf emphasize the semantic consistency of salient postures}, i.e., postures with large movement scope that often correspond to strong semantics. We design a weakly-supervised salient posture detector to identify salient postures, %, and contribution of the crucial pose representation to the audio-to-gesture mapping during the joint training process. %More concretely, we take the pose sequence of body part as input to train the detector, which is only under the weak supervision of the sequence-level salient label. We develop 
which utilizes a temporal relation module to mine long-range temporal dependencies among the pose features and %project the pose features into the 1D vector space to obtain the 
predicts frame-level saliency score. 
We then use the saliency score to reweight the consistency loss %along the frame axis. The integration of saliency score can effectively facilitate the model to focus more on learning the mapping relationship between the complicated postures and corresponding audios with the highly semantic information.
to enforce a stronger alignment between the salient postures and corresponding audios in the joint embedding.

In addition, facial expressions, especially lip motions, mainly rely on articulation-related acoustic features; while body gestures are closely correlated with strong semantics in the speech content.
%Therefore, we employ an auxiliary face synthesis branch to leverage the rhythmic representation of audio to generate the face part with more synchronized expression. 
Therefore, we extract audio features dedicated to facial expressions and body gestures separately, and synthesize facial expressions and body gestures with separate branches.
Meanwhile, we enforce the temporal alignment between the audio features extracted in the face branch and the body branch, which effectively improves synchronization and naturalness between the face and body parts. Our main contributions can be summarized as follows:

1) We propose a novel speech-driven gesture generation framework with an emphasis on semantic consistency of salient posture. We introduce the joint manifold space to learn the inherent semantic association between audio and gesture modalities and enforce semantic consistency via a consistency loss.

2) We emphasize the semantic consistency of salient postures by introducing a weakly-supervised detector to identify salient postures, and reweighting the consistency loss based on saliency score to enforce a stronger alignment in the joint manifold space.% detector to identify the contributions of different levels of poses to the joint mapping, which facilitates the model to learn the correspondence between salient poses and the high-level semantic information of audios.

3) Observing that facial expressions rely on articulation-related audio features while body gestures rely on semantic-related features, we propose to extract separate audio features for face and body, and design separate branches for face and body synthesis. %an auxiliary generation branch to synthesize more synchronized facial expressions. 

\section{Related Work}
% \subsection{Audio-Visual Cross-Modal Learning} 
\noindent\textbf{Audio-Visual Cross-Modal Learning.} Multi-modal machine learning aims to train models capable of processing and relating information from multiple modalities. 
Recent works~\cite{ahuja2019language2pose,ahuja2020style,yoon2020speech} encode all the modalities into a common representation space. 
Language2Pose~\cite{ahuja2019language2pose} learns a joint embedding of text and pose. 
Ahuja \textit{et al.}~\cite{ahuja2020style} maps the learned style embedding along with audio into a joint gesture space. 
Trimodal~\cite{yoon2020speech} utilizes separate representations for different modalities and handles the alignment between speech and gesture explicitly. 
Compared with these methods, our method focuses on identifying semantic correlations between audio and gestures to synthesize more natural-looking and vivid gesture sequences.
% }

\begin{figure*}[t]
    \centering
    \includegraphics[scale=0.5]{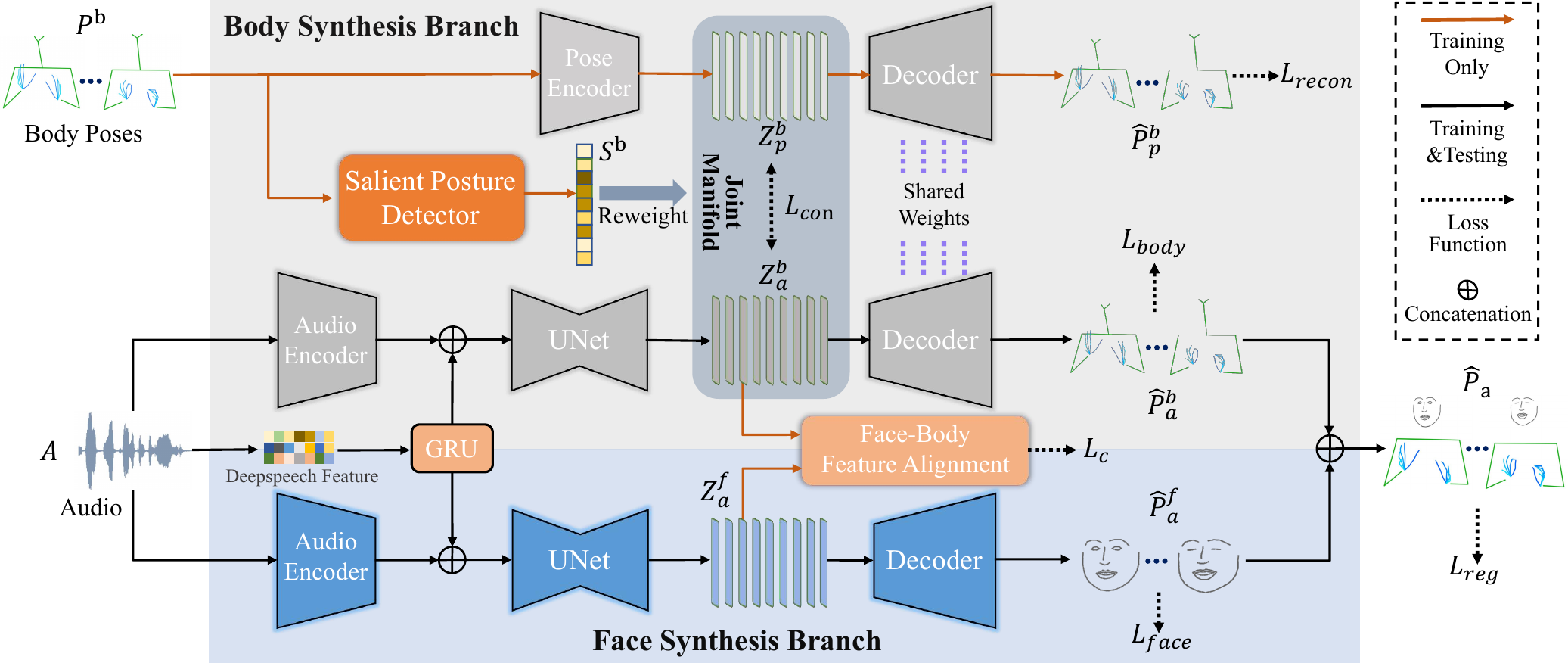}
    \caption{The overall architecture of our proposed method, which consists of two branches including body synthesis branch and face synthesis branch. In the body synthesis branch, our model learns a joint manifold space of representations to enforce semantic consistency by employing a dual-path structure, which contains the upper reconstruction path and the lower speech-driven generation path. Furthermore, the salient posture detector is designed to identify salient gestures and reweight the consistency loss. We then generate synchronized facial expressions using face synthesis branch. Finally, we fuse the generated results of two branches to obtain the entire gesture sequence. }
    \label{fig:framework}
\end{figure*}

% \subsection{Speech-Driven Gesture Generation} 
\noindent\textbf{Speech-Driven Gesture Generation.}
Synthesizing consistent and natural gestures in line with speech is becoming popular in the field of multi-modal generation. 
% Modalities involved in the gesture generation task can be divided into three categories: driving modalities such as speech audio, target modalities like gesture sequences, and middle modalities such as style and identity embeddings.
% {\color{blue}Early rule-based methods~\cite{kopp2006towards,wagner2014gesture,cassell1994animated} suffer from the fact that audio-gesture correlation is too complex to be learned directly and fail to generate realistic results.}
With the development of deep learning, recent works~\cite{shlizerman2018audio,ginosar2019learning,alexanderson2020style,qian2021speech,xu2021exploring,gong2024demos,fan2024freemotion} leverage deep neural networks to generate more natural motion and gesture sequences. 
% Learning the mapping from audio to visual gesture mainly faces the challenge of multi-modal prediction ambiguity. To tackle the problem,
Audio2Body~\cite{shlizerman2018audio} learns the correlation between audio features and body landmarks through an LSTM network and concentrates on predicting body motion from specific music like piano and violin.
S2G~\cite{ginosar2019learning} incorporates generative adversarial learning into the regression-based prediction model to enhance the naturalness of generated results. 
MoGlow~\cite{alexanderson2020style} adapts the probabilistic model to this task to learn the distribution of gesture motions and can effectively take control over gesture styles. 
SDT~\cite{qian2021speech} learns template vectors to provide extra information for gesture generation and transform the one-to-many ambiguous regression problem into a deterministic conditional regression problem. 
However, these methods ignore the semantic association of different modalities and fail to deal with habitual and salient gestures.
Our method utilizes joint manifold space to model the mapping function and design a salient posture detector to maintain semantic consistency of salient gestures.

% \subsection{Anomaly Detection} 
\noindent \textbf{Anomaly Detection.}
Video anomaly detection aims to identify abnormal events in the video that do not match the normal behaviors. 
Early works~\cite{mehran2009abnormal,hospedales2009markov,kim2009observe,ClaudioPiciarelli2008TrajectoryBasedAE} focus on detecting anomalies through manually extracting features and modeling the anomalies. 
% With the advancement of deep neural network techniques, 
Recently, several deep learning-based methods~\cite{hasan2016learning,luo2017revisit,YiruZhao2017SpatioTemporalAF,liu2018future,lu2019future}  have achieved significant performance improvement. 
% Such methods can be divided into three types: reconstruction-based, prediction-based, and the hybrid of both. 
Reconstruction-based methods~\cite{hasan2016learning,luo2017revisit,YiruZhao2017SpatioTemporalAF} utilize the autoencoder trained on normal datasets and detect anomalous frames which are difficult to be reconstructed well. 
Prediction-based methods~\cite{liu2018future,lu2019future} predict future frames and utilize prediction error as an indicator to determine anomalies, considering the frames with large prediction errors as anomalies. 
% Hybrid approaches~\cite{MuchaoYe2019AnoPCNVA,TrongNguyenNguyen2019AnomalyDI} further combine two types of methods to obtain better reliability.
% {\color{blue}
Different from these methods, we design a weakly-supervised salient posture detector to identify anomaly gestures only under the weak supervision of video-level labels.
and reweight the cross-modal association based on the predicted saliency score. 
% }

\section{Method}
\subsection{Overview and Notations}
To fully empower the learning of semantic association between speech and gesture, we propose a novel speech-driven gesture synthesis method that emphasizes semantic consistency of salient postures, i.e., postures with large movement scope that often correspond to strong semantics. Our overall architecture is shown in Figure~\ref{fig:framework}. 
Our model first learns a joint manifold space for different representations of audio and body pose to explore a finer mapping between two modalities. 
Then, %we take the pose sequence of body as input to train the salient posture detector, which can identify the contribution of the crucial pose representation to the audio-to-gesture mapping during the joint training process. 
we emphasize the semantic consistency of salient postures by introducing a weakly-supervised detector to identify salient postures, and enforcing a stronger alignment for the salient postures in the joint manifold space.
In addition, %we propose an auxiliary face synthesis branch to leverage the rhythmic feature of audio to generate facial expressions with better synchrony. 
observing that facial expressions rely on articulation-related audio features while body gestures rely on semantic-related audio features, we extract separate audio features for face and body, and design separate branches for face and body gesture synthesis.

\begin{figure*}[t]
    \centering
    \includegraphics[scale=0.7]{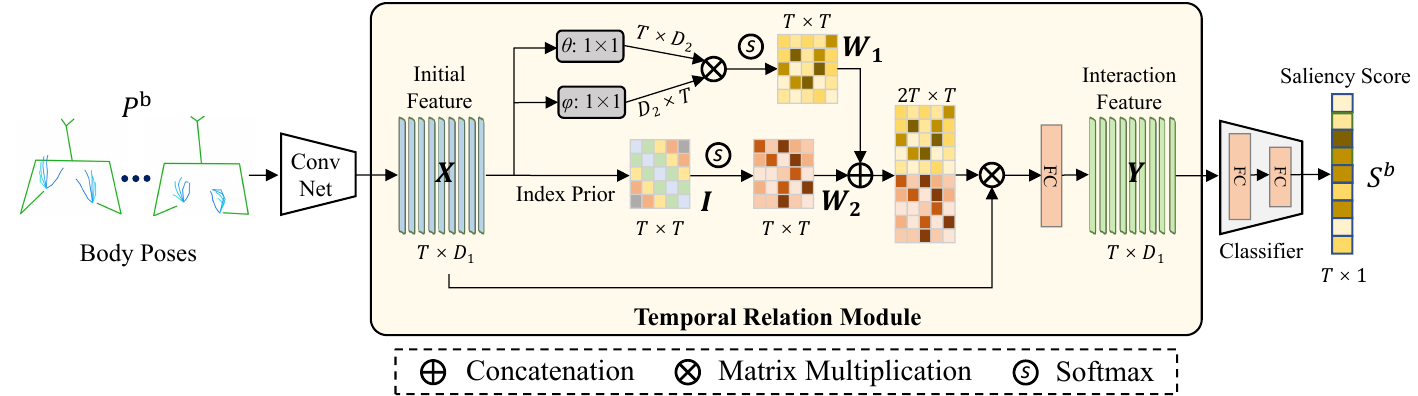}
    \caption{The detailed structure of proposed salient posture detector. We take as input the real body poses $P^b$ and extract the initial feature $X$ using the ConvNet. Then, $X$ is fed into the temporal relation module to obtain the interaction feature $Y$. We utilize a classifier to map $Y$ to the 1D salient score $S^b$, which is used to reweight the consistency loss of joint manifold training.}
    \label{fig:detector}
\end{figure*}
% Finally, the body branch and face branch are 
% \subsection{Preliminary}
Given an audio clip $A = \bracks{A_1, \dots, A_{T}}$, speech-driven gesture synthesis aims to generate a pose sequence $P = \bracks{P_1, \dots, P_{T}}$ synchronized with the audio, where the length of sequence is $T$ and $P_t$ is the pose at the $\raisedth{t}$ frame. Each $P_t$ is defined as the coordinates set of $J$ upper-body keypoints in a frame, including face, hands, and arms. 
The major challenge of synthesizing realistic gestures is how to establish the reasonable mapping: $A_{1:T} \to P_{1:T}$. Due to the discrepancy of mapping property of different parts, we decompose the entire pose $P_t$ into body part $P^b_t$ and face part $P^f_t$. The whole gesture sequence $P$ can be formulated as: $P = P^b \oplus P^f$.
% \begin{equation}
%     P = P^b \oplus P^f.
% \end{equation} 

% More specifically, 
For the generation of body poses, we first utilize two feature extractors to extract audio and pose latent codes respectively. Then, we use a pose decoder to separately generate poses from these two latent codes. The output of our pose decoder is a sequence of $T$ frames. We denote the pose sequence generated from pose latent codes as $\hat{P}^b_p$, and the pose sequence generated from audio latent codes as $\hat{P}^b_a$. Then, for facial expressions synthesis, we extract different audio features (articulation-related) and directly generate a sequence of $T$ frames $\hat{P}^f_a$. Eventually, by fusing the synthesis results of the two branches, we obtain the final output of the entire gesture sequence $\hat{P} = \hat{P}_a = \hat{P}^b_a \oplus \hat{P}^f_a$.

% in the upper part of  Figure~\ref{fig:framework}

\subsection{Joint Manifold Space for Speech and Gesture}
Due to the high randomness of body motion, our model learns a multi-modal joint manifold space between audio and body pose to explore the semantic correlation between audio and pose representations. To form the shared joint-embedding space, as shown in the upper part of Figure~\ref{fig:framework}, we employ a dual-path architecture that consists of two parallel pipelines: the reconstruction path for body pose and the speech-driven gesture generation path. The reconstruction path takes as input the real pose sequence of body part $P^b$, and uses a pose encoder $Enc_p$ consisting of two GRUs to obtain a pose feature in the joint manifold space, 
\begin{equation}
    Z^b_p = Enc_p(P^b),
\end{equation}
where $Z^b_p \in \mathbb{R}^{T \times D}$ is the latent code of body pose and $D$ represents the dimension of the latent code.

For the speech-driven gesture generation path, we first utilize the audio encoder $Enc_a$ to encode the mel-spectrogram of audio and concatenate it with the feature code $f_d$ extracted by the $DeepSpeech$~\cite{amodei2016deep} model, \lfq{which is pre-trained by large numbers of audio-transcript pairs. The integration of high-dimensional representations of $DeepSpeech$ can provide richer semantic information for follow-up gesture generation.}
% deepspeech model 
We further map the concatenated feature to an audio feature in the joint manifold space using 1D UNet translation network,
\begin{equation}
    Z^b_a = UNet(Enc_a(A) \oplus f_d),
\end{equation}
where $Z^b_a \in \mathbb{R}^{T \times D}$ is the latent code of audio corresponding to body part. 

% first derive a 2D mel-spectrogram feature from the original audio through frequency transformation following previous works~\cite{ginosar2019learning,qian2021speech}. 
% For clarity, we continue to denote the mel-spectrogram feature by $A$.
To guarantee that $Z^b_p$ and $Z^b_a$ lie close to each other in the shared joint-embedding space and represent similar semantic information, we propose the consistency loss to constrain the latent codes. Similar to the cosine similarity function, the consistency loss calculates the cosine similarity between the body pose latent code and audio latent code:
\begin{equation}
    \label{eq:alignment loss}
    L_{con} = \sum_{t=1}^T (1-\frac{Z_{p,t}^b \cdot Z^b_{a,t}}{\max (|| Z_{p,t}^b||\cdot ||Z^b_{a,t} ||,\epsilon)}),
\end{equation}
where $Z_{p,t}^b$ and $Z^b_{a,t}$ are the $\raisedth{t}$ value of latent codes along time axis respectively, and $\epsilon$ is a fairly small positive scalar. 

Then, the decoders of the reconstruction path and generation path learn to generate the pose sequences $\hat{P}^b_p$ and $\hat{P}^b_a$ from the above two latent codes,
\begin{equation}
   \hat{P}^b_p = Dec(Z^b_p), \quad \hat{P}^b_a = Dec(Z^b_a).
\end{equation}
Here, two decoders share the same network parameters and both are denoted as $Dec$.

\subsection{Weakly-supervised Salient Posture Detector}
% {\color{blue}
Salient postures are the postures with a large range of motion, which are closely related to strong semantics of the speech content. 
% e.g., the movements in which the shoulders and arms of speakers have a relatively large amplitude of motion, and closely related to strong semantics of the speech content. 
%, and also closely related to the habitual actions of the speaker. 
% It is critical for generating vivid and realistic co-speech gestures to focus more on the salient and personalized postures.
Therefore, we propose a weakly-supervised salient posture detector to predict the saliency score of poses for each frame, which will then be used to reweight the consistency loss. 
This strategy enforces a more accurate alignment of salient pose representation and corresponding audio representation in the joint embedding. 
The architecture of the detector is depicted in Figure~\ref{fig:detector}. 
First, we utilize the inflated ConvNet~\cite{carreira2017quo} to extract the initial frame-level feature $X \in \mathbb{R}^{T \times D_1}$, where $D_1$ denotes the dimension of the initial feature. Then, we feed $X$ into a temporal relation module to capture long-range temporal dependencies among features and transform $X$ into the interaction feature $Y \in \mathbb{R}^{T \times D_1}$. Then, a classifier is used to project the interaction feature into the 1D vector space to obtain the frame-level saliency score of poses $S^b \in \mathbb{R}^T$. We use the average saliency score of $top$-$k$ frames %over the temporal dimension 
as the saliency prediction for the video sequence, and make it close to the video-sequence-level saliency label using the binary cross-entropy loss.

 % (video sequences containing salient postures are assigned label $1$ and otherwise $0$)
 
% {\color{blue}
\subsubsection{Sequence-level Salient Label}
To facilitate the network to adaptively learn which frames are salient, we train the detector under the weak supervision of the video-sequence-level saliency labels. To acquire the sequence-level labels, we first derive the resting pose (the most frequent posture) of each speaker in the entire dataset as illustrated in the S2G ~\cite{ginosar2019learning}. For a pose sequence, we calculate the $L_{2}$ distance between the pose of each frame and the resting pose. Then we can obtain the probability distribution of all distances and determine whether the sequence contains salient data points based on the triple variance criterion. Video sequences containing salient points are assigned label $1$ and otherwise $0$.

Additionally, the ground truth of poses is pseudo and obtained using OpenPose~\cite{ZheCao2018OpenPoseRM}, therefore it is inaccurate to directly take the saliency points as the frame-level labels. Sequence-level saliency labels can mitigate this error and facilitate the detector and generation network to learn the exact correspondence
between salient postures and speech semantics. See the comparison results in Table~\ref{tab: ablation_label} of the ablation study.
% }

% }

% \subsubsection{\textbf{Temporal Relation Module.}}
% \noindent \textbf{Temporal Relation Module.}
\subsubsection{Temporal Relation Module.}

Inspired by the temporal attention architecture~\cite{wu2021learning},  we design the temporal relation module (TRM) to mine the most relevant information between frames to facilitate salient posture detection. Compared with ~\cite{wu2021learning}, our TRM leverages the global information of a sequence instead of local information in neighbor frames to exploit long-range temporal dependencies, and capture the index prior of all frames to improve the representation capability. 

Specifically, we first compute the adjacency matrix for the initial frame-level feature $X$ to measure the affinity in the embedding space, and then normalize the matrix using softmax function along each row. We denote the normalized adjacency matrix as weight matrix $W_1$, which can be formulated as:
\begin{equation}
    W_1 = softmax(\theta(X)^\top \varphi (X)),
\end{equation}
where $\theta$ and $\varphi$ denote linear transformation functions. 

Then we incorporate the frame index prior into the network to distinguish the importance of neighbor frames to the current frame. 
% {\color{blue}
We denote the index prior matrix as $I = \bracks{i_{1}, i_{2}, \dots,i_{t}, \dots,i_{T}}$, where $i_t$
% $i_t= \bracks{-(t-1),\dots,-1,0,-1,\dots,-(T-t)}$ 
is the negative relative distance vector between the current frame ($\raisedth{t}$) and all frames. 
% }
Since frames closer to the current frame have more relevant information, we use the softmax function to transform the index matrix $I$ into another weight matrix $W_2$, where closer frames are mapped to higher weights.
We then concatenate $W_2$ with $W_1$ to compute the final interaction feature $Y$:
\begin{equation}
    Y = \phi((W_1 \oplus W_2) \times X ),
\end{equation}
where $\phi$ denotes the transformations implemented by FC layers.

\subsubsection{Emphasizing Consistency for Salient Posture.}

%Here, we utilize a classifier including two linear layers to map the interaction feature $Y$ to the 1D vector $S^b$, which is regarded as the salient score of body pose in the entire sequence. 
After we obtain the frame-level saliency score for body poses, we use the saliency score to reweight the consistency loss in Eq.(\ref{eq:alignment loss}) to enforce stronger semantic consistency for salient postures (as shown in the upper part of Figure~\ref{fig:framework}). The reweighted consistency loss can be re-formulated as follows:
\begin{equation}
    L_{con} = \sum_{t=1}^T S^b_t \cdot (1-\frac{Z_{p,t}^b \cdot Z^b_{a,t}}{\max (|| Z_{p,t}^b||\cdot ||Z^b_{a,t} ||,\epsilon)}),
\end{equation}
where $S^b_t$ is the predicted saliency score for the body pose in the $t$-th frame.
The reweighted consistency loss with saliency score can effectively enforce the model to focus more on learning the mapping between salient pose and the high-level semantics of audio during the joint training.

% \subsection{Full Face-Body Synthesis}
\subsection{Separate Face and Body Synthesis}
% \subsubsection{Motivation for Separate Synthesis.}

\subsubsection{Separate Audio Feature for Face and Body.}
% Previous works~\cite{qian2021speech, JingXu2022FreeformBM} generate face and body gestures as a whole with single branch of networks. We found that 
Synthesizing facial expressions and body gestures as a whole often leads to poor lip synchronization, since different audio features are required for the generation of face and body parts: articulation-related acoustic features are required for face and semantic-related acoustic features for body. 
% The prediction of facial expressions, especially for lip motions, mainly relies on articulation-related acoustic features, while the large salient movements of the body parts are closely correlated with strong semantics of audio. 
% Based on the above observation, 
Therefore, we extract audio features dedicated to facial expressions and body gestures separately, and synthesize facial expressions and body gestures with separate branches. As shown in the lower part of Figure~\ref{fig:framework}, for the face synthesis branch, we utilize a separate audio encoder and UNet to extract the rhythmic representation of audio $Z^f_a \in \mathbb{R}^{T \times D}$. Then, we decode $Z^f_a$ to generate a sequence of $T$ frames of facial keypoints $\hat{P}^f_a$ using a decoder, which has the same network structure as the body synthesis branch but different parameters. 
% Specifically, the audio encoder consists of several 2D convolutional down-sampling layers, the encoder and decoder of U-net are implemented as 1D convolutional layers with skip connections. The final decoder is composed of several 1D convolutional layers to convert the channel numbers to $J_f \times 2$ and obtain the face pose result.
% In the face branch, we also take deepspeech audio feature and face part of seed pose as input to enhance the generating quality. 
% The incorporation of the separate face synthesis branch effectively facilitates our model to generate diverse facial expressions with better lip-synchronization.

\begin{figure}[t]
    \centering
    \includegraphics[width=\linewidth]{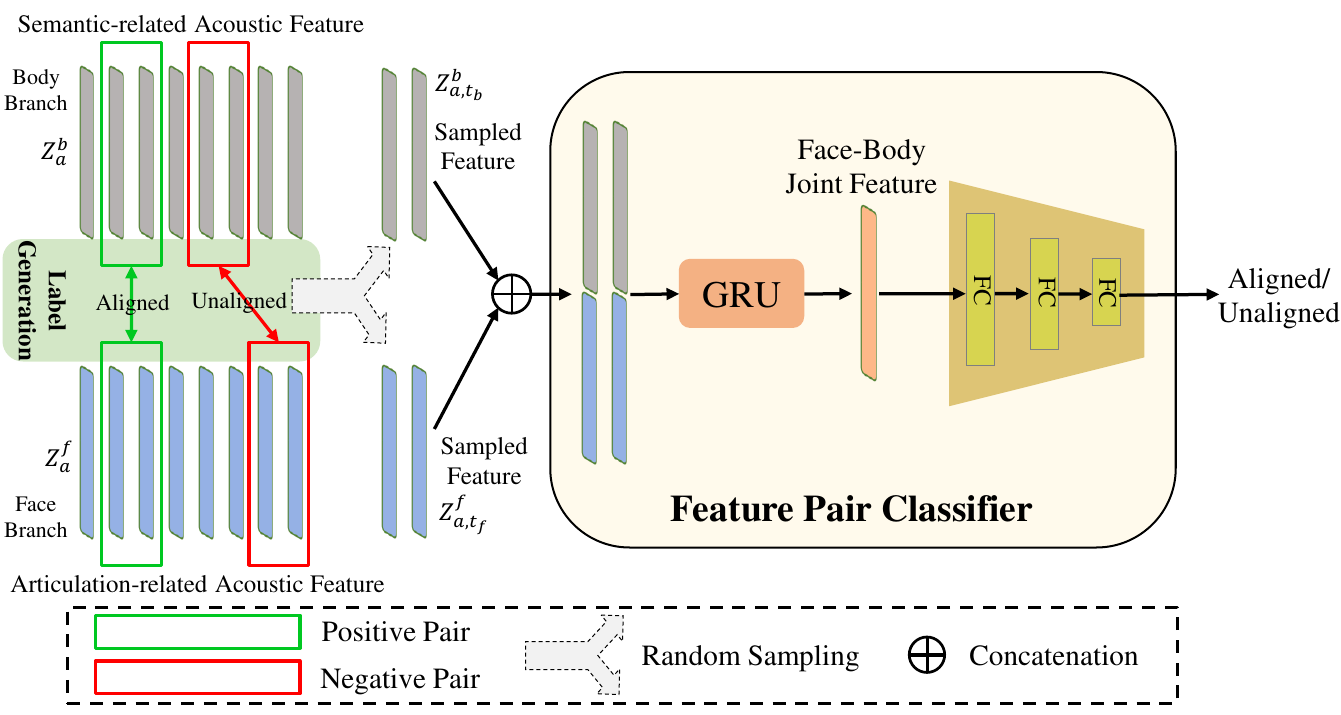}
    \caption{The detailed structure of the face-body feature alignment module, which is trained by a self-supervised manner.}
    \label{fig:ssl}
\end{figure}
\subsubsection{Face-Body Feature Alignment.}

\begin{table*}[thb]
	\centering
	\renewcommand{\tabcolsep}{1.2mm}
	\small
         \caption{Quantitative comparison of all methods on four speakers (Oliver and Kubinec are from the S2G \cite{ginosar2019learning} dataset, Xing and Luo are collected by SDT~\cite{qian2021speech}).  We use $L_2$ dist. and FGD metrics to evaluate the accuracy and realism of generated results. BC and PSD metrics are utilized to evaluate the synchronization. 
	 }
        \label{tab:2d dataset}
	\begin{tabular}{l|cccc|cccc|cccc|cccc}
		\toprule
            \multirow{2}{*}{Methods}
            &\multicolumn{4}{c|}{Oliver} &\multicolumn{4}{c|}{Kubinec} &\multicolumn{4}{c|}{Xing} &\multicolumn{4}{c}{Luo} \\ \cline{2-17}
		&\multicolumn{1}{c}{$L_2$ dist. $\downarrow$} 
		&\multicolumn{1}{c}{FGD$\downarrow$} &\multicolumn{1}{c}{BC$\uparrow$} 
		&\multicolumn{1}{c|}{PSD$\downarrow$}
		&\multicolumn{1}{c}{$L_2$ dist. $\downarrow$} 
		&\multicolumn{1}{c}{FGD$\downarrow$} &\multicolumn{1}{c}{BC$\uparrow$} 
		&\multicolumn{1}{c|}{PSD$\downarrow$}
		&\multicolumn{1}{c}{$L_2$ dist. $\downarrow$} 
		&\multicolumn{1}{c}{FGD$\downarrow$} &\multicolumn{1}{c}{BC$\uparrow$} 
		&\multicolumn{1}{c|}{PSD$\downarrow$}
		&\multicolumn{1}{c}{$L_2$ dist. $\downarrow$} 
		&\multicolumn{1}{c}{FGD$\downarrow$} &\multicolumn{1}{c}{BC$\uparrow$} 
		&\multicolumn{1}{c}{PSD$\downarrow$}\\ 
		\midrule
		Audio2Body~\cite{shlizerman2018audio}
		&49.7  &3.48  &0.27  &6.29
            &70.9  &4.51  &0.23  &6.20			
            &50.9  &4.75  &0.19  &6.41  
            &\textbf{48.4}  &2.70  &0.21  &7.81 \\
		S2G~\cite{ginosar2019learning} 
		&53.5  &8.30  &0.63  &5.97  
            &64.9	&4.53  &0.61  &6.33		
            &48.0  &4.49  &0.65  &6.25	 
            &63.7	&3.10  &0.57  &8.12  \\
		MoGlow \cite{alexanderson2020style} 
		&50.6  &2.28  &0.32  &6.23  
            &78.1  &2.49  &0.29  &6.39			
            &48.4  &4.94  &0.35  &6.34	 
            &54.8	&1.47  &0.30  &7.86  \\
		SDT~\cite{qian2021speech}
		&62.4  &0.92  &0.63  &6.12  
            &100.7  &1.07  &0.65  &6.37	
            &57.8  &1.72  &0.68  &6.52  
            &80.8	 &0.69  &0.59  &7.97  \\ 
            SEEG~\cite{liang2022seeg} 
            &52.9  &0.83  &0.55  &6.58  
            &63.6  &1.60  &0.53  & 5.98
            &47.1  &1.89  &0.56  &6.40  
            &62.4	 &0.82  &0.49  &7.24  \\ 
            DiffGesture~\cite{zhu2023taming}
            &48.2  &0.76  &0.71  &6.01 
            &57.0  &1.33  &0.68  & 5.72
            &40.6  &1.95  &0.73  &5.98  
            &50.0	 &0.75  &0.64  &7.33  \\ 
            Ours   
            &\textbf{35.7}  &\textbf{0.55}  &\textbf{0.78}  &\textbf{5.83}
            &\textbf{42.6}  &\textbf{0.46}  &\textbf{0.72}  &\textbf{5.69}
            &\textbf{37.9}  &\textbf{1.56}  &\textbf{0.76}  &\textbf{5.91}
            & 51.4  &\textbf{0.49}  &\textbf{0.68}  &\textbf{6.95}  \\
		\bottomrule
	\end{tabular}
\end{table*}

\begin{table}[thb]
	\centering
	\renewcommand{\tabcolsep}{1.2mm}
	\small
         \caption{Quantitative results on TED Expressive dataset. We compare the proposed method with other recent methods under FGD, BC, and PSD metrics. 
	 }
        \label{tab:3d dataset}
	\begin{tabular}{l|ccc}
		\toprule
            \multirow{2}{*}{Methods}
            &\multicolumn{3}{c}{TED Expressive} \\ \cline{2-4}
		&\multicolumn{1}{c}{FGD$\downarrow$} &\multicolumn{1}{c}{BC$\uparrow$}  &\multicolumn{1}{c}{PSD $\downarrow$}\\ 
		\midrule
		Attention Seq2Seq~\cite{yoon2019robots}
		&54.92  &0.15  & 7.58 \\
		S2G~\cite{ginosar2019learning} 
		&54.65  &0.68  & 7.95  \\
		Joint Embedding \cite{ahuja2019language2pose} 
		&64.56  &0.13  &  7.52 \\
		Trimodal~\cite{yoon2020speech}
		&12.61  &0.56   & 7.89  \\ 
            HA2G~\cite{liu2022learning} 
            &5.31  &0.64   &  7.11 \\ 
            DiffGesture~\cite{zhu2023taming}
            & 2.60 & \textbf{0.72}   & 6.97  \\
            Ours 
            & \textbf{2.50} & 0.68 & \textbf{6.83} \\
		\bottomrule
	\end{tabular}
\end{table}

Although the separate body and face branches can yield individual prediction results well aligned with audio, the final gesture sequences generated by direct concatenation of the two results tend to be inconsistent and unnatural. 
% {\color{blue}
% Since the synchronization between audios and generated results of both branches is not necessarily the same, 
Therefore, we design a binary classification task at the feature level to enforce the temporal alignment between representations of the body branch and face branch.
% }
 % , inspired by ~\cite{MunroJonathan2019MultiModalDA}. 
As shown in Figure~\ref{fig:ssl}, from the semantic-related acoustic feature $Z^b_a$ (for body gesture synthesis) and the articulation-related acoustic feature $Z^f_a$ (for facial expression synthesis), we randomly sample body feature sequence $Z^b_{a,t_b}$ and face feature sequence $Z^f_{a,t_f}$, which have the same length of frames $T_c$. Here, $t_b$ and $t_f$ represent the starting index of the feature sequences respectively. We
 denote the one-hot label as $c \in \mathbb{R}^2$ to distinguish whether the audio features corresponding to the body and face parts are well aligned over the temporal dimension. Then, we can obtain the labeled feature pairs including the aligned positive pairs ( if $t_b=t_f$, then $c=[1,0]$ ) and the unaligned negative pairs ( if $t_b \neq t_f$, then $c=[0,1]$ ). We take the labeled feature pairs as training data to optimize the binary classifier $C$ to determine whether the two input features are temporally aligned. As depicted in Figure~\ref{fig:ssl}, the structure of $C$ consists of a GRU and three FC layers.
% We design the architecture of the classifier as shallow as possible to guarantee that the self-supervised branch features are mostly learned by the audio encoder and U-net. Considering that the input of the classifier is two feature sequences, we first use a single-layer GRU to extract the timing-related information, which is then processed by a 3-layer MLP to obtain the final classification result. 
With this module, we can align the feature space of both branches by self-supervised learning, resulting in realistic gesture generation with better synchrony. Formally, the loss function of the classification task can be formulated as:
\begin{equation}
    L_{c} = \sum_{t_b,t_f \in [1,T-T_c]} -c\log C(Z^b_{a,t_b}, Z^f_{a,t_f}),
\end{equation}
where $C(*,*)$ is the classification result of classifier $C$.

\subsection{Objective Function}
\textbf{Reconstruction Loss.}
The goal of our reconstruction path of the body synthesis branch is to learn the pose representation space by recovering the input body pose sequence. Here, we use $L_1$ loss function to measure the distance between the reconstructed body poses $\hat{P}^b_p$ and ground truth pose sequences $P^b$:
\begin{equation}
    L_{recon} = \frac{1}{T} \sum_{t=1}^T || \hat{P}^b_{p,t} - P^b_t ||_1.
\end{equation}

\noindent \textbf{Regression Loss.}
The main supervision of the training process is imposed on the pose keypoints, including the supervision of the overall keypoints as well as the individual part keypoints. 
The supervision on the overall pose keypoints is implemented as a $L_1$ regression loss between the entire gestures $\hat{P}_a$ generated from the given audio and ground truth gestures $P$:
\begin{equation}
    L_{reg} = \frac{1}{T} \sum_{t=1}^T || \hat{P}_{a,t}-P_t ||_1.
\end{equation}
Besides, we additionally apply separate constraints on face pose keypoints and body pose keypoints by measuring the distance between gestures $\hat{P}^b_a, \hat{P}^f_a$ generated from the given audio, and ground truth $P^b, P^f$ with Huber loss ($HL$)~\cite{huber1992robust}:
\begin{equation}
    \begin{aligned}
    L_{body}  =\sum_{t=1}^T HL(\hat{P}_{a,t}^b,P_t^b),  \quad
    L_{face}  =\sum_{t=1}^T HL(\hat{P}_{a,t}^f,P_t^f).
    \end{aligned}
\end{equation}
Then the entire loss of face part and body part is calculated as the mean value of $L_{face}$ and $L_{body}$:
\begin{equation}
    L_{huber} = \frac{1}{2T}  (L_{body}+L_{face}).
\end{equation}

Overall, the total training objective function is:%for the whole framework
%can be computed as follows:
\begin{equation}
\begin{aligned}
    L_{total} &= \lambda_{r} L_{recon} + \lambda_{reg} L_{reg} + \lambda_{h} L_{huber} \\      
    & + \lambda_{con} L_{con} + \lambda_{c} L_{c},
\end{aligned}
\end{equation}
where $\lambda_{r},\lambda_{reg},\lambda_{h},\lambda_{con}$, and $\lambda_{c}$ are hyper-parameters that can be adjusted to control the relative significance of each loss term.

% dynamically 

\section{Experiments}
\subsection{Datasets}
\textbf{Speech2Gesture.}
Speech2Gesture~\cite{ginosar2019learning} is a speaker-specific dataset with full body and face keypoints annotations. Following SDT~\cite{qian2021speech}, we test our method on four speakers: Oliver, Kubinec, Luo, and Xing. 
The number of different videos of four speakers is 113, 274, 72, and 27, with a total length of about 25 hours. 
All pseudo keypoint annotations are obtained by OpenPose~\cite{ZheCao2018OpenPoseRM}, which contains $121$ keypoints. We divide them into different parts, including $70$ keypoints for the facial expression, $51 $ keypoints for the body pose. 

\noindent\textbf{TED Expressive.} TED Expressive dataset~\cite{liu2022learning} is derived from a large-scale 3D gesture dataset TED Gesture~\cite{yoon2020speech}. Compared to TED Gesture only with 10 upper body key points, TED Expressive is more realistic and expressive of both body and finger movements. Through 3D pose estimator ExPose~\cite{choutas2020monocular} extracting the pose information, TED Expressive contains the 3D coordinate annotations of 43 keypoints, including 13 upper body joints and 30 finger joints.

\begin{figure*}[t]
    \centering
    \includegraphics[width=0.8\linewidth]{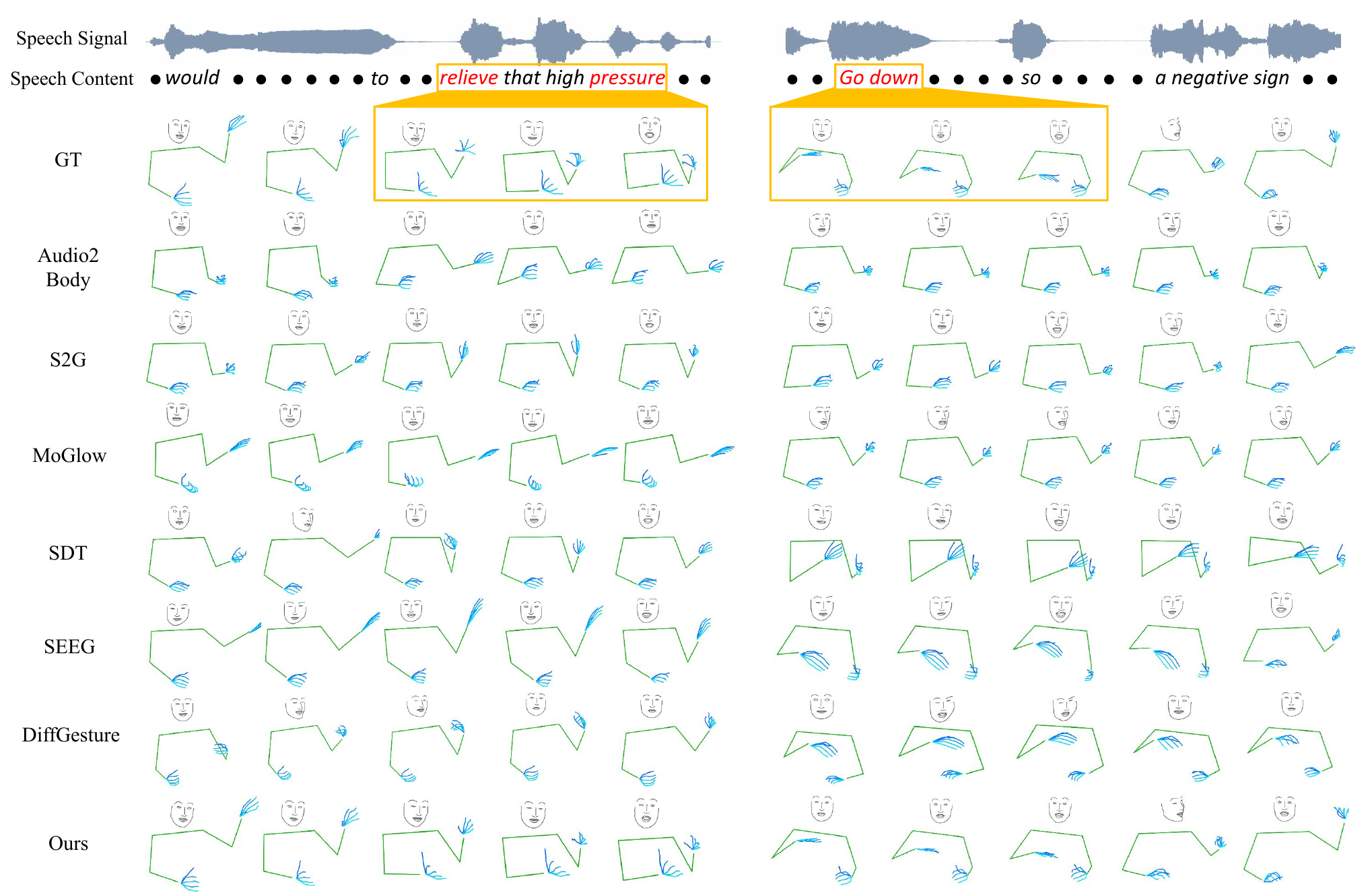}
    \caption{Visualization results of generated gesture sequence of all methods given the speech signal. Our method can synthesize more natural and realistic gestures with better synchrony than others.  }
    \label{fig:quaComp}
\end{figure*}

\subsection{Implementation Details}
\label{exp:settings}

We use the data pre-processing protocol in SDT~\cite{qian2021speech} to partition videos of $15$ FPS into segments with $64$ frames for training. 
We use a 1-layer GRU as our pose feature extractor with the hidden size of $1024$ and the dimension $D$ of our audio-pose joint embedding space is $512$. For the salient posture detector module, the dimension $D_1$ and $D_2$ of initial feature and interaction feature are set to 512 and 1024. In addition, we set the value of $top$-$k$ to 16. For both training and testing, we use a batch size of $32$. We train our model with an Adam optimizer and the learning rate is set to $0.0001$. 
For the hyper-parameters , we empirically set $\lambda_{r}=10, \lambda_{reg}=10, \lambda_{h}=20, \lambda_{con}=1$, and $\lambda_{c}=1$, which work well for both datasets.

\subsection{Quantitative Evaluation}
\subsubsection{Evaluation Metrics.}
% We employ the following four metrics to quantitatively evaluate the generation performance of all methods. \\
% in Sec.~\ref{exp:settings}. \\
$\bm{L_2}$ \textbf{Distance} is commonly used to measure the distance between the generated gestures and ground truth. \\
\textbf{Fr\'{e}chet Gesture Distance (FGD)} is proposed by ~\cite{yoon2020speech} to measure the distribution distance of ground truth gestures and generated ones in the latent space. Note that we use the same pose feature extractor as ~\cite{qian2021speech} for a fair comparison. \\
% \textbf{Lip-Sync Error (LSE)} is originally proposed by ~\cite{qian2021speech} to measure the synchronization between audio and lip motion. They denote lip-sync error as the average distance between the center keypoints of the upper and lower lip. 
\textbf{Beat Consistency Score (BC)} is originally proposed by ~\cite{liu2022learning,zhu2023taming} to measure the beat correlation between audio and gesture. They utilize the angle changes of bones to quantify the motion beat and calculate the average distance between audio beat and its nearest motion beat as Beat Consistency Score.  
\\
\textbf{Pose-Sync Distance (PSD)} is designed by us to evaluate the consistency of audio and generated pose sequence. 
Inspired by SyncNet~\cite{JoonSonChung2016OutOT}, we train a Pose-SyncNet using a similar contrastive loss between the representations of audio and generated pose. We compute this evaluation metric using the $L_2$ distance between the audio embedding and pose embedding of our pre-trained Pose-SyncNet. Concretely, for a pose sequence $\mathbf{p}$ of $64$ frames with corresponding audio $\mathbf{a}$, we evenly divide the pose sequence into $N$ pose clips $p_i$ of $9$ frames and audio clips $a_i$ of $0.6s$, and encode them using Pose-SyncNet to obtain corresponding pose feature $f_{p_i}$ and audio feature $f_{a_i}$. Then, we calculate the PSD metric as follows:
\begin{equation}
    PSD = \frac{1}{N} \sum_{(p_i,a_i)\in (\mathbf{p},\mathbf{a})} || f_{p_i} - f_{a_i} ||_2.
\end{equation}

\subsubsection{Evaluation Results.}
We make a comprehensive performance comparison between our method and other state-of-the-art methods on both datasets using the above metrics.
As shown in Table \ref{tab:2d dataset} and Table~\ref{tab:3d dataset}, our method almost achieves the highest performance in all evaluation metrics, which demonstrates great superiority over existing methods. 
In terms of $L_2$ distance and FGD metrics, our method outperforms Audio2Body~\cite{shlizerman2018audio}, S2G~\cite{ginosar2019learning}, and MoGlow~\cite{alexanderson2020style} by a large margin, indicating that our method can produce more accurate prediction results while better maintaining the realism and diversity of the generated gestures. The incorporation of joint-embedding space and salient gesture detection can effectively enforce the semantic consistency between audio and gesture, which is critical for generating vivid and realistic co-speech gestures. Compared with SDT~\cite{qian2021speech}, SEEG~\cite{liang2022seeg}, and HA2G~\cite{liu2022learning}, our method can generate gestures with better synchronization for both the lip motions and body pose movements, resulting in the higher BC and lower PSD. This is due to the separate audio feature extraction and synthesis branch for facial expressions and body gestures. 
Overall speaking, with the combination of the joint manifold training and separate synthesizing, our method generates the most plausible co-speech gestures and achieves a better balance between synchronization and versatility. 

\begin{table}
\renewcommand{\arraystretch}{1.15}
\centering
\small
\caption{
User study results of the gesture sequences generated by different methods on naturalness, synchronization, and expressiveness. The rating scores range from highest (7) to lowest (1).
}
\label{table:user study}
% \resizebox{\linewidth}{!}{
\begin{tabular}{l|c|c|c}
\toprule
 Methods&Expressiveness & Naturalness & Synchronization  \\
\midrule
GT & 6.56 & 6.62 & 6.37 \\
Audio2Body~\cite{shlizerman2018audio} & 1.18 & 2.88 & 1.34 \\
S2G~\cite{ginosar2019learning}  & 2.74 & 1.65 & 3.82 \\ 
MoGlow~\cite{alexanderson2020style} & 4.16 & 3.90 & 2.26 \\
SDT~\cite{qian2021speech} & 3.45 & 3.25 & 3.05  \\
SEEG~\cite{liang2022seeg} &4.02 & 4.72 &  4.30 \\
DiffGesture~\cite{zhu2023taming} &5.15 & 5.36 &  5.04 \\
\midrule
Ours & 5.89 & 6.01 & 6.12 \\
\bottomrule
\end{tabular}

\end{table}

\subsection{Qualitative Evaluation}
\subsubsection{Visualization Analysis.}
% {\color{blue}
As shown in Fig.~\ref{fig:quaComp}, we visualize the generated gestures of all methods given the speech signal and compare our method with other methods. 
In the left case of Fig. \ref{fig:quaComp}, when the speaker says the verb phrase \textit{relieve pressure}, he folds his open arms down to the lower right, making a gathering movement to express the meaning of the phrase.
Only our method successfully synthesizes this strong semantic-related gesture since we emphasize the semantic consistency of salient posture in our method.
% by designing a weakly-supervised detector to identify the salient posture. 
The naturalness of gestures generated by Audio2Body~\cite{shlizerman2018audio}, S2G~\cite{ginosar2019learning} and SEEG~\cite{liang2022seeg} is not good enough, especially the realism of hand motion generated by S2G~\cite{ginosar2019learning} is visually poor and the face shape generated by SEEG~\cite{liang2022seeg} is distorted and deformed.
MoGlow~\cite{henter2020moglow}, SDT~\cite{qian2021speech}, and DiffGesture~\cite{zhu2023taming} generate less diversity in the poses. 
In addition, for the right case of Fig. \ref{fig:quaComp}, the speaker lowers his raised right hand when he says the phrase \textit{go down}. Audio2Body~\cite{shlizerman2018audio} and MoGlow~\cite{henter2020moglow} generate gestures of similar appearance with a small range of motion and unnatural hands. 
% The template of SDT~\cite{qian2021speech} restricts it from generating poses with large variation, thus failing to learn strong semantic gestures. 
SEEG~\cite{liang2022seeg} and DiffGesture~\cite{zhu2023taming} generate the coarse pose appearance, but the generated right hand is sagging when the one of ground truth is flat. Compared with these methods, our approach can learn the salient posture and generate more realistic results.

\subsubsection{User Study.}
We conduct a subjective user study to compare our method with other baselines from three aspects: naturalness, synchronization, and expressiveness. 
We collect anonymous gesture videos of all methods for four characters in the dataset and invite 25 volunteers to watch and score them based on the three metrics described above. The rating scores range from $1$ to $7$, with $7$ being the most plausible and $1$ being the least plausible. As shown in Table~\ref{table:user study}, our method shows significant advantages over other approaches and even achieves comparable results against ground truth. Compared with Audio2Body~\cite{shlizerman2018audio} and S2G~\cite{ginosar2019learning}, our method performs better on the gesture naturalness and synchronization due to the separate face-body synthesis framework. In addition, the integration of salient posture detection enhances the expressiveness of our method, which outperforms MoGlow~\cite{alexanderson2020style}, SDT~\cite{qian2021speech}, SEEG~\cite{liang2022seeg}, and DiffGesture~\cite{zhu2023taming} by a large margin.

\begin{table}[t]
\renewcommand{\arraystretch}{1.15}
\centering
\small
\caption{Effectiveness of key components of our framework. \textit{Separate}, \textit{Joint}, and \textit{Detector} respectively denote separate synthesis, joint manifold space, and salient posture detector.}
\label{tab: key_components}
\resizebox{0.99\linewidth}{!}{
\begin{tabular}{c|ccc|ccc}
\toprule
Baseline & \textit{Separate}  & \textit{Joint}   & \textit{Detector} & FGD  $\downarrow$  & BC $\uparrow$  & PSD $\downarrow$  \\
\midrule
\cmark &  &   &   & 4.27 & 0.53 &  6.56  \\
\cmark & \cmark &  &   & 3.95 & 0.68 &  6.33  \\
\cmark & \cmark & \cmark  &  & 1.78 & 0.71 & 6.02 \\
\cmark &\cmark & \cmark  & \cmark  & \textbf{0.46} & \textbf{0.72} & \textbf{5.69} \\
\bottomrule
\end{tabular}}
\end{table}

\begin{table}[t]
\renewcommand{\arraystretch}{1.15}
\centering
\small
\caption{Impact of different salient label strategies.}
\label{tab: ablation_label}
\begin{tabular}{l|ccc|ccc}
% \toprule
% Methods & FGD  $\downarrow$  & LSE $\downarrow$  & PSD $\downarrow$ & FGD  $\downarrow$  & LSE $\downarrow$  & PSD $\downarrow$  \\
% \midrule

\toprule
\multirow{2}{*}{Strategy}
 &\multicolumn{3}{c|}{Oliver} &\multicolumn{3}{c}{Kubinec}\\ \cline{2-7}
&\multicolumn{1}{c}{FGD $\downarrow$ } &\multicolumn{1}{c}{BC $\uparrow$ } 
&\multicolumn{1}{c|}{PSD $\downarrow$ }
&\multicolumn{1}{c}{FGD $\downarrow$ } &\multicolumn{1}{c}{BC $\uparrow$ } 
&\multicolumn{1}{c}{PSD $\downarrow$ }
\\ 
\midrule

Frame-level & 1.94  & 0.73  & 6.30 & 2.01 & 0.70 & 6.25 \\
Squence-level  & \textbf{0.55} & \textbf{0.78} & \textbf{5.83}  & \textbf{0.46} & \textbf{0.72} & \textbf{5.69} \\

\bottomrule
\end{tabular}
\end{table}

\subsection{Ablation Study}

We conduct extensive ablation studies to justify the contribution of key components to the final performance of the proposed method.

\noindent \textbf{Effectiveness of key components.} Table~\ref{tab: key_components} summarizes the performance and effectiveness of different design components on the speaker of Kubinec. We use the baseline model for a fair comparison with other variants of our method, which only contains an audio encoder and pose decoder and directly predicts the holistic gesture sequence using the audio representation. From the results of the first and second row in Table~\ref{tab: key_components}, we can see that compared with the baseline, the integration of separate synthesis for facial expressions can facilitate the significant improvement of synchronization (See the obvious increase of BC metric). In addition, our model learns a joint manifold space to exploit the inherent semantic association between speech content and gesture, which helps to achieve the lower FGD metric and effectively enhance the realism of generated gestures. 
As shown in the last two rows in Table~\ref{tab: key_components}, the incorporation of the salient detector module can further decrease the FGD metric and achieve an obvious performance boost in the PSD metric.

\noindent \textbf{Impact of different salient label strategies.}
We report the performance of different salient label strategies using the same baseline on Oliver and Kubinec in Table~\ref{tab: ablation_label}. Frame-level strategy means training the salient posture detector under the supervision of frame-level salient labels. We acquire the frame-level labels according to the distances between the poses of all frames in a sequence and the resting pose. As shown in Table~\ref{tab: ablation_label}, we observe that compared to the frame-level strategy, the proposed sequence-level approach can boost performance, especially in terms of FGD and PSD metrics. This is because sequence-level weak supervision facilitates the detector and generation network to learn the exact correspondence between salient postures and speech content.

\section{Conclusion}
In this paper, we propose a novel co-speech gesture generation method to enhance the learning of cross-modal association of speech and gesture. Our model learns a joint manifold space for different representations of audio and body pose to exploit the inherent association between two modalities and enforce semantic consistency using a consistency loss. 
% {\color{blue}
Further, we introduce a weakly-supervised salient posture detector to facilitate the model to focus more on learning the mapping of salient postures and corresponding audios with highly semantic information. 
Extensive experiments demonstrate that the proposed method can effectively enhance the naturalness and fidelity of generated gestures.

\begin{acks}
This work was supported by National Natural Science Foundation of China (72192821, 62302297, 62272447), Shanghai Sailing Program (22YF1420300), Shanghai Municipal Science and Technology Major Project (2021SHZDZX0102), Shanghai Science and Technology Commision (21511101200), Young Elite Scientists Sponsorship Program by CAST (2022QNRC001), Beijing Natural Science Foundation (L222117), the Fundamental Research Funds for the Central Universities (YG2023QNB17, YG2024QNA44).
\end{acks}

{
    \small
    \bibliographystyle{ACM-Reference-Format}
    \bibliography{main}
}

\end{document}